\documentclass{article}

\usepackage{microtype}
\usepackage{graphicx}
\usepackage{subcaption}
\usepackage{xspace}
\usepackage{booktabs} 

\usepackage{hyperref}

\usepackage[accepted]{icml2026}

\usepackage{amsmath}
\usepackage{amssymb}
\usepackage{mathtools}
\usepackage{amsthm}

\usepackage[capitalize,noabbrev]{cleveref}

\theoremstyle{plain}

\theoremstyle{definition}

\theoremstyle{remark}

\usepackage[textsize=tiny]{todonotes}

\icmltitlerunning{When One Adapter Speaks for Many: Discovering Low-Rank Redundancy in Continual Fine-Tuning}

\begin{document}

\newcommand{\our}[0]{\textsc{LiteLoRA}\xspace}

\twocolumn[
  \icmltitle{When One Adapter Speaks for Many: Discovering Low-Rank Redundancy in Continual Fine-Tuning}



  \icmlsetsymbol{equal}{*}

  \begin{icmlauthorlist}
    \icmlauthor{Tanguy Dieudonné}{eth}
    \icmlauthor{Giulia Lanzillotta}{eth}
    \icmlauthor{Enis Simsar}{eth}
    \icmlauthor{Louis Barinka}{eth}
    \icmlauthor{Thomas Hofmann}{eth}
  \end{icmlauthorlist}

  \icmlaffiliation{eth}{Department of Computer Science, ETH Zurich, Zurich, Switzerland}

  \icmlcorrespondingauthor{Tanguy Dieudonné}{tdieudonne@ethz.ch}

  \icmlkeywords{Continual Learning, LoRA, Low-Rank Adaptation, Lifelong Learning}

  \vskip 0.3in
]



\printAffiliationsAndNotice{}  

\begin{abstract}
Low-Rank Adaptation (LoRA) has become the standard tool for parameter-efficient fine-tuning of large pretrained models. When applied sequentially across tasks in Continual Learning (CL), the standard assumption is that each new task requires a dedicated low-rank adapter. In this work, we challenge this assumption empirically and structurally. We show that task-specific LoRA adapters in CL exhibit significant \emph{low-rank redundancy}: the subspaces spanned by adapters trained on different tasks substantially overlap, and in many cases earlier adapters can faithfully represent later tasks. Building on this observation, we propose \textbf{\our}, a \textit{plug-and-play} gating mechanism that learns at train time whether to recruit a new adapter or reuse existing low-rank representations. Our method reduces the number of active adapters by 20–70\% while matching or exceeding state-of-the-art performance on standard CL benchmarks, revealing that structural redundancy is pervasive and that selective learning is sufficient to achieve stability without sacrificing plasticity.
\end{abstract}

\section{Introduction}

In the era of foundation models, the need for effective and \emph{efficient} adaptation is ever growing. As pretraining costs scale dramatically with model size, repeatedly retraining models from scratch to incorporate new data is increasingly impractical. This motivates a shift toward \emph{lifecycle-aware adaptation}, where models are incrementally updated over time. Parameter Efficient Fine Tuning (PEFT) have emerged as a key mechanism for enabling such updates at low cost. Instead of updating all backbone weights, PEFT methods introduce lightweight modules, such as low-rank adapters (LoRA) \citep{hu2021loralowrankadaptationlarge}, allowing foundation models to acquire new capabilities while limiting computational and memory overhead.

Despite their reduction in costs, PEFT methods still face a fundamental challenge in sequential fine-tuning settings: simply acquiring new information may overwrite existing knowledge and degrade overall model performance, a phenomenon called \emph{Catastrophic Forgetting} (CF).

Continual Learning (CL) algorithms \citep{HADSELL20201028} address CF by ensuring the \emph{stability} of past knowledge during adaptation.
In the context of Low-Rank Adaptation, the prevalent strategy is to impose active constraints on adapters to minimize cross-task interference. For instance, O-LoRA \citep{wang2023orthogonal} orthogonalizes adapters across tasks, while InfLoRA \citep{liang2024inflora} aligns the adapter's row space with the task input span. While effective at preserving stability, these constraints inevitably reduce the model's \emph{plasticity}---its ability to learn new tasks---as the available optimization space becomes increasingly restricted over time.
Conversely, recently \citet{wu2025sdlora} showed that decoupling the magnitude and direction of updates reduces forgetting without restricting the parameter space, proposing a constraint-free method named SD-LoRA. While this preserves plasticity, the lack of constraints can lead to unbounded forgetting as the task sequence grows. Consequently, existing methods force a binary choice between rigidity (high stability, low plasticity) and drift (high plasticity, low stability).

Our work is motivated by a simple empirical question: do different tasks in a CL stream require genuinely distinct low-rank subspaces, or do they share latent structure? We show empirically that the answer is strongly the latter. We propose \textbf{\our}, a method that learns a differentiable binary gate to determine, for each new task, whether existing adapters already span the relevant low-rank subspace or whether a new one is needed. Built on top of SD-LoRA \citep{wu2025sdlora}, \our \space uses 20--70\% fewer adapters while matching or exceeding state-of-the-art performance.

\section{Problem Setup}

\paragraph{PEFT and LoRA.} Parameter-efficient fine-tuning via low-rank adaptation (LoRA) is a widely adopted approach for fine-tuning pretrained models to downstream tasks. Specifically, given a pretrained model with weight matrix $\mathbf{W}_0 \in \mathbb{R}^{m \times n}$, LoRA \citep{hu2021loralowrankadaptationlarge} introduces the parameter update $\Delta \mathbf{W} = \mathbf{AB} \in \mathbb{R}^{m \times n}$ where $\mathbf{A} \in \mathbb{R}^{m \times r}$ and $\mathbf{B} \in \mathbb{R}^{r \times n}$ are two low-rank learnable matrices with rank $r \ll \min\{m,n\}$. The LoRA-updated output becomes $\boldsymbol{h} = (\mathbf{W}_0 + \mathbf{AB})\boldsymbol{x}$. 

\paragraph{Continual Learning.} In CL, a sequence of tasks $\{\mathcal{T}_1, \cdots, \mathcal{T}_T \}$ must be learned sequentially. The \textit{t}-th task $\mathcal{T}_t$ consists of a training dataset $\mathcal{D}_t = \{x_t^{(i)}, y_t^{(i)} \}_{i=1}^{|\mathcal{D}_t|}$ where $x_t^{(i)}$ denotes an input image and $y_t^{(i)}$ its corresponding label. When training on $\mathcal{D}_t$, the model does not have access to previous data $\{\mathcal{D}_k\}_{k=1}^{t-1}$ but is required to mitigate catastrophic forgetting of the $t-1$ previously learned tasks. We follow the class-incremental learning setting where the model must classify all classes seen across all tasks, without being provided the task identity at inference.
When applying LoRA to CL, during learning on task $\mathcal{T}_t$, we compute the output at a given layer as
\[
\boldsymbol{h} = (\mathbf{W}_0 + \mathbf{A}_1 \mathbf{B}_1 + \mathbf{A}_2 \mathbf{B}_2 + \cdots + \mathbf{A}_t \mathbf{B}_t)\boldsymbol{x}.
\]

\section{\our}
\label{method}

We propose a modular gating framework that can wrap any adapter-based CL method.

\subsection{Learnable Gating Mechanism}
Given a sequence of tasks, each task $i$ is associated with a candidate adapter 
$\Delta\mathbf{W}_i \in \mathbb{R}^{m \times n}$. The cumulative weight update 
after learning task $t$ is:
\begin{equation}
\label{eq:update}
\mathbf{U}_t = \sum_{i=1}^{t} \beta_{i} \cdot \Delta\mathbf{W}_i,
\end{equation}
where $\Delta\mathbf{W}_i$ is the adapter for task $i$ and $\beta_i \in \{0,1\}$ 
is a binary gate determining whether it contributes to the forward pass. For previously learned tasks ($i < t$), $\beta_i$ is fixed after training. For the current task, we initialize $\beta_t \equiv 1$ to learn candidate features, and subsequently optimize it to determine whether the newly introduced adapter is necessary.

We employ a \emph{global gating strategy}, where a single scalar logit $l_i$ controls the activation of the adapter for task $i$ across all layers. This formulation encourages the model to make a binary decision at the task level: either the new task requires a dedicated adapter, or it can be solved by reusing existing knowledge.

\paragraph{Differentiable Selection via Gumbel-Sigmoid.}
To make the binary decision differentiable, we parameterize $\beta$ using a continuous logit $l$. We employ the Gumbel-Sigmoid relaxation \citep{jang2016categorical} combined with the Straight-Through Estimator (STE) \citep{DBLP:journals/corr/BengioLC13}.
During the forward pass, we compute a ``soft'' gate $\beta_{\text{soft}}$ and discretize it:
\begin{equation}
\label{eq:gumbel_sigmoid}
\beta_{\text{soft}} = \sigma\left(\frac{l + G}{\tau}\right), \quad \beta = \mathbb{I}(\beta_{\text{soft}} > 0.5),
\end{equation}
where $\sigma$ is the sigmoid function, $\tau$ is a temperature parameter (fixed to $\tau=1$), and $G \sim \text{Gumbel}(0,1)$ is noise sampled during training.
To enable gradient flow, we use the STE formulation:
\begin{equation}
\label{eq:ste}
\beta_{\text{train}} = \beta - \mathrm{SG}(\beta_{\text{soft}}) + \beta_{\text{soft}},
\end{equation}
where $\mathrm{SG}(\cdot)$ denotes the stop-gradient operator. This allows the network to use discrete binary weights in the forward pass while gradients propagate through the continuous $\beta_{\text{soft}}$ in the backward pass.
At \emph{training time} we add \emph{Gumbel noise} $G$ to the new logit, as we found that this would lead to increased levels of sparsity. Intuitively, the added noise may randomly turn on and off the adapter during training, acting like a dropout regularizer. Importantly, \emph{the Gumbel noise is sampled only during training}. At inference time, we disable the noise and compute a deterministic selection using the learned logit $\beta = \mathbb{I}(\sigma(l_i) > 0.5)$. This ensures reproducible predictions while the training process benefits from stochastic exploration. We find that Gumbel noise acts as a dropout-like regularizer; see Appendix~\ref{sec:noise_ablation_appendix} for an ablation, but remark that it is not essential to the method. By default, noise is injected only in Phase~2; Phase~1 is deterministic (the ``No phase 1 noise (Ours)'' row in Table~\ref{tab:comparison_results}).

\paragraph{Two-phase Training}

Simultaneously optimizing the adapter parameters ($\mathbf{A}, \mathbf{B}$) and the structural gate ($l$) can lead to instability. Therefore, we decouple the learning process for each task $t$ into two distinct phases:

\begin{itemize}
    \item \textbf{Phase 1: Feature Acquisition.} We freeze all previous adapters and the current gate logit $l_t$ (initializing $\beta_t=1$). We train the new adapter parameters $\{\mathbf{A}_t, \mathbf{B}_t, \alpha_t\}$ and the classification head. This phase focuses entirely on learning the features necessary to solve the new task, assuming the capacity is available.
    \item \textbf{Phase 2: Structural Pruning.} This phase is extremely lightweight: one epoch over a single scalar versus thousands of adapter parameters in Phase 1. We freeze the newly learned adapter parameters and the classification head, and train \emph{only} the scalar logit $l_t$ under a global sparsity penalty (the full training loss is given in Eq.~\ref{eq:total_loss}, Appendix~\ref{sec:hyperparams}). Since all previously learned gating logits $\{l_i\}_{i < t}$ are frozen, the sparsity penalty on their corresponding gates is constant and does not contribute to the gradients. As a result, only the current logit $l_t$ is updated.
    This phase forces the model to verify if the new adapter is strictly necessary. If the task can be solved using the accumulated knowledge of previous adapters, the sparsity penalty drives $\beta_t$ to 0.
\end{itemize}

At the end of Phase 2, we evaluate the final gate $\beta_t$. If $\beta_t=0$, the parameters $\mathbf{A}_t, \mathbf{B}_t$ are permanently discarded, resulting in zero parameter growth for that task.

\section{Empirical analysis}
\label{sec:experiments}

\paragraph{Benchmarks.} We evaluate \our \space on three benchmark datasets for class-incremental learning. \emph{CIFAR-100}~\citep{cifar} contains 100 classes of $32{\times}32$ images, which we split into 10 disjoint tasks of 10 classes each. \emph{ImageNet-A}~\citep{hendrycks2021nae} consists of 200 classes of natural adversarial examples; we split it into 20 disjoint tasks of 10 classes each. \emph{ImageNet-R} ~\citep{hendrycks2021many} consists of 200 ImageNet classes ~\citep{imagenet} rendered in artistic styles; we split it into 20 tasks of 10 classes each.   

\paragraph{Implementation.} We instantiate \our \space on top of SD-LoRA \citep{wu2025sdlora}, using its magnitude--direction decomposition $\Delta\mathbf{W}_i = \alpha_i \cdot \overline{\mathbf{A}_i\mathbf{B}_i}$ as the per-task adapter, chosen for its strong plasticity properties, where $\overline{\mathbf{A}\mathbf{B}} = \frac{\mathbf{A}\mathbf{B}}{\|\mathbf{A}\mathbf{B}\|_F}$.
The gating mechanism is otherwise backbone-agnostic and could equally wrap O-LoRA or InfLoRA adapters. We adopt the setup of SD-LoRA \citep{wu2025sdlora}, which has been shown to outperform prior LoRA-based CL methods such as InfLoRA \citep{liang2024inflora}. For all three datasets we use a pre-trained ViT-B/16 backbone \citep{DBLP:conf/iclr/DosovitskiyB0WZ21} with a hidden dimension $d=768$. LoRA adapters of rank $r=10$ are applied to the Query and Value projections in all 12 Transformer blocks.

\paragraph{Metrics.} We evaluate performance using two standard metrics: \textbf{Average Anytime Accuracy (A)}, defined as the mean accuracy across all seen tasks evaluated after each incremental step; and \textbf{Forgetting (F)}, which measures the average decline in performance for each task relative to its peak value.

\subsection{\our \space matches accuracy with far fewer adapters}

As detailed in \cref{tab:combined_results}, our method consistently matches or exceeds the accuracy of SD-LoRA while substantially reducing parameter growth. In particular, our method has two important characteristics:
\begin{itemize}
    \item \textbf{Efficiency:} On CIFAR-100, we achieve comparable accuracy using only 5--8 adapters, compared to the fixed 10 used by SD-LoRA. This efficiency gain is even more pronounced on the two 20-task ImageNet benchmarks: on ImageNet-A (Order 1), our method requires only 6 adapters versus 20 for the baseline, and on ImageNet-R it uses just 6--7 adapters across all three orderings, corresponding to a $65$--$70\%$ reduction in added parameters.
    \item \textbf{Robustness:} Despite the variation in active adapter count across orderings and datasets, accuracy remains stable. This demonstrates that the gating mechanism successfully identifies the minimal capacity required for each curriculum, maintaining a strong accuracy--efficiency trade-off even under both adversarial-style shift (ImageNet-A) and rendition/style shift (ImageNet-R).
\end{itemize}

\subsection{The ordering matters for efficiency}

Whether a new adapter is recruited depends on the semantic gap between the new task and those already learned: if existing representations suffice, phase 2 discards the new adapter. The order in which tasks are visited therefore directly controls total adapter count. Across three orderings on all three benchmarks (\Cref{tab:combined_results}), the number of active adapters varies from 5 to 8 on CIFAR-100, from 6 to 14 on ImageNet-A, and from 6 to 7 on ImageNet-R, while accuracy remains stable throughout, confirming that the additional adapters are redundant rather than necessary. On ImageNet-A Order 1, retaining just 6 adapters improves average accuracy by 0.9 points over SD-LoRA, suggesting this ordering presents tasks in a sequence where prior representations transfer well. ImageNet-R exhibits an even more stable pruning profile across orderings, which is consistent with the idea that rendition shifts preserve reusable structure that can be captured by a small shared set of low-rank adapters. In all cases, the gating mechanism adapts to the curriculum.

\begin{table*}[t]
\centering

\caption{\textbf{Main Results.} Comparison of SD-LoRA and our method (\our) on CIFAR-100 ($N=10$), ImageNet-A ($N=20$), and ImageNet-R ($N=20$) across three different task orderings. We report Average Accuracy (A$\uparrow$), Forgetting (F$\downarrow$), and the number of active adapters (\#Ad$\downarrow$).}
\small
\setlength{\tabcolsep}{3pt}
\begin{tabular}{lccccccccc}
\toprule
& \multicolumn{3}{c}{\textbf{Order 1}}
& \multicolumn{3}{c}{\textbf{Order 2}}
& \multicolumn{3}{c}{\textbf{Order 3}} \\
\cmidrule(lr){2-4} \cmidrule(lr){5-7} \cmidrule(lr){8-10}
Method & A$\uparrow$ & F$\downarrow$ & \#Ad$\downarrow$
       & A$\uparrow$ & F$\downarrow$ & \#Ad$\downarrow$
       & A$\uparrow$ & F$\downarrow$ & \#Ad$\downarrow$ \\
\midrule
\multicolumn{10}{l}{\textit{CIFAR-100 ($N=10$)}} \\
\midrule
SD-LoRA
    & 91.71 & 5.92 & 10
    & 92.54 & \textbf{4.93} & 10
    & 91.27 & \textbf{5.96} & 10 \\
Ours
    & \textbf{91.81} & \textbf{4.66} & \textbf{5} 
    & \textbf{92.64} & 5.14 & \textbf{8} 
    & \textbf{91.41} & 6.09 & \textbf{7} \\ 
\midrule
\midrule
\multicolumn{10}{l}{\textit{ImageNet-A ($N=20$)}} \\
\midrule
SD-LoRA
    & 64.85 & 18.53 & 20
    & 66.30 & \textbf{14.70} & 20
    & 62.08 & \textbf{15.14} & 20 \\
Ours
    & \textbf{65.75} & \textbf{16.55} & \textbf{6} 
    & \textbf{66.53} & 17.27 & \textbf{14} 
    & \textbf{63.64} & 16.49 & \textbf{13} \\ 
\midrule
\midrule
\multicolumn{10}{l}{\textit{ImageNet-R ($N=20$)}} \\
\midrule
SD-LoRA
    & 82.41 & 10.25 & 20
    & 81.55 & 8.89 & 20
    & 81.27 & 13.67 & 20 \\
Ours
    & \textbf{82.83} & \textbf{7.84} & \textbf{6} 
    & \textbf{81.64} & \textbf{8.49} & \textbf{7} 
    & \textbf{81.71} & \textbf{12.26} & \textbf{6} \\ 
\bottomrule
\end{tabular}
\label{tab:combined_results}
\end{table*}

\subsection{Low-Rank Redundancy is Structured}

\begin{figure}[h]
    \begin{center} \includegraphics[width=1.0\linewidth]{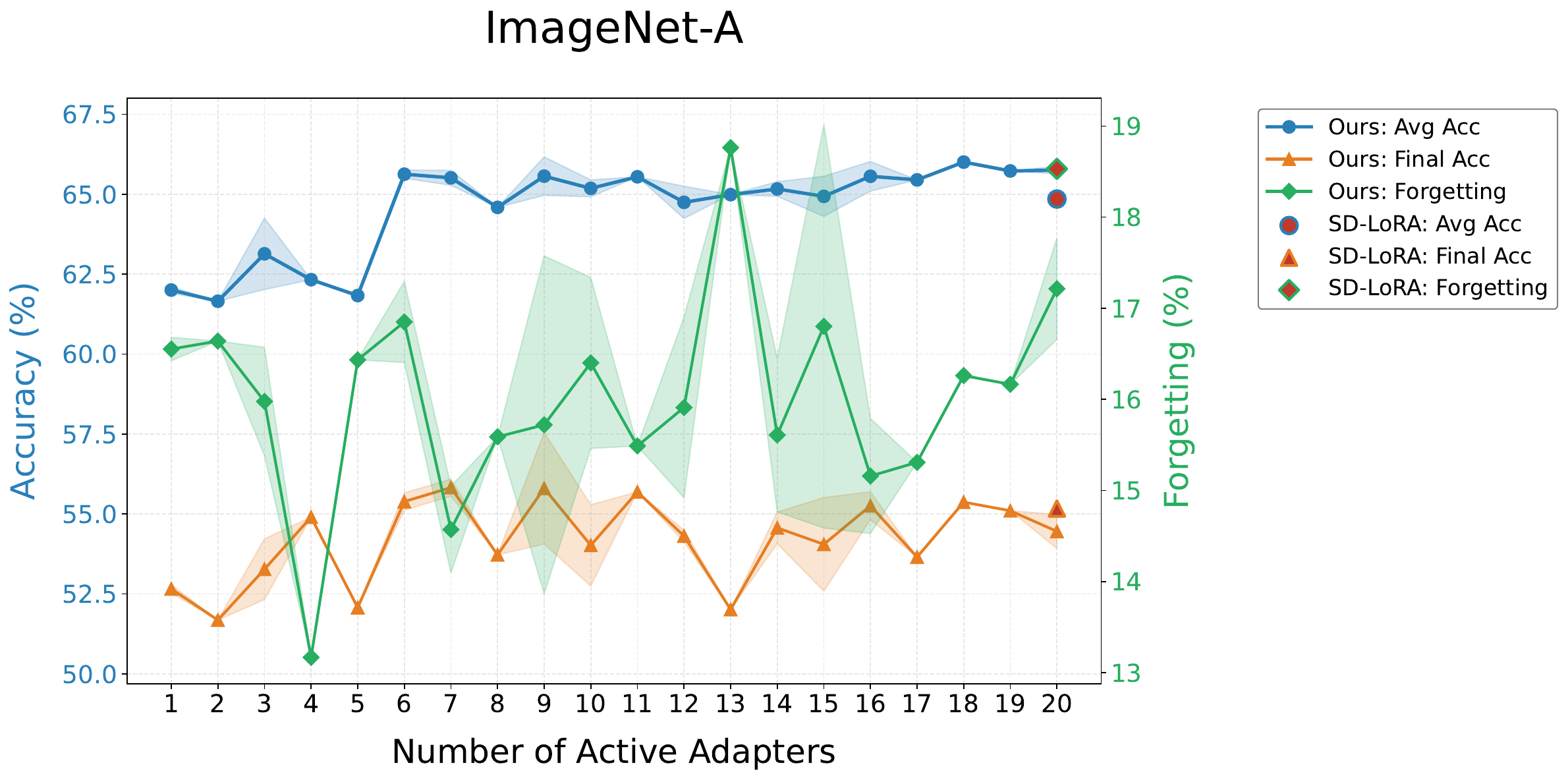}
    \caption{Sparsity--accuracy frontier on ImageNet-A (order 1). We report average accuracy and forgetting as a function of the number of active adapters.}
    \label{fig:sparsity_frontier_ina}
    \end{center}
\end{figure}

We now ask not just how many adapters are needed, but whether the redundancy itself has stable structure. To answer this, we sweep the sparsity regularization strength $\lambda_\text{sparsity}$ across a broad range, yielding configurations with active adapter counts varying from $1$ to the full sequence length $N$.

Figure~\ref{fig:sparsity_frontier_ina} illustrates the resulting sparsity--accuracy frontier on ImageNet-A; the CIFAR-100 frontier is given in Appendix~\ref{sec:pruning_consistency}. We observe that average accuracy saturates quickly, often peaking at intermediate values rather than at the full budget.
For instance, on CIFAR-100, configurations with 5--7 active adapters outperform the SD-LoRA baseline by $0.1\%$ and our own dense (10-adapter) model by $0.3\%$. Similarly, on ImageNet-A, the 6-adapter configuration surpasses SD-LoRA by $0.9\%$ and remains within $0.13\%$ of our 20-adapter model, despite a $70\%$ reduction in parameters. These results establish that the low-rank subspace needed to represent a stream of $N$ tasks is spanned by far fewer than $N$ adapters.

This trend confirms our driving hypothesis: excessive adapter capacity increases the risk of forgetting without improving plasticity. As sparsity increases, forgetting generally decreases while accuracy remains stable. However, extreme sparsity (e.g., 1--2 adapters) eventually degrades performance, highlighting the necessity of our data-driven selection rule over a fixed, minimal budget.

Further evidence for this comes from sweeping $\lambda_{\text{sparsity}}$: on CIFAR-100 (Table~\ref{tab:pruning_order_cifar}), the pruned set grows monotonically with $\lambda$, revealing a stable importance hierarchy. On ImageNet-A (Appendix~\ref{sec:pruning_consistency}) the ordering is noisier, but Tasks 5 and 8 are pruned in 88\% of configurations, confirming that the gating mechanism captures meaningful redundancy structure even across 20 fine-grained tasks. Together, these patterns \textbf{validate} the hypothesis that task-level gating learns a meaningful redundancy structure, promoting the selective reuse of previously learned low-rank components.

\begin{table}[h]
\centering
\caption{\textbf{CIFAR-100 Pruning Order.} Evolution of the pruned task set as the sparsity penalty $\lambda$ increases. The strictly nested sets (each row is a superset of the previous) indicate a stable hierarchy of task importance. Minor deviations occur at $\lambda{=}0.04$ and $\lambda{=}0.05$ due to the stochastic nature of the Gumbel gate. Task~6 is consistently the first to be removed,
suggesting its knowledge is already well-captured by earlier adapters.}
\label{tab:pruning_order_cifar}
\begin{tabular}{ccl}
\toprule
$\lambda_{\text{sparsity}}$ & Active & Pruned Tasks (Cumulative) \\
\midrule
$\leq 0.01$ & 10 & $\emptyset$ \\
$0.015$ & 9 & $\{6\}$ \\
$0.02$ & 8 & $\{4, 6\}$ \\
$0.027$ & 7 & $\{2, 4, 6\}$ \\
$0.028$ & 6 & $\{2, 4, 6, 9\}$ \\
$0.04$ & 5 & $\{2, 4, 6, 7, 8\}$ \\
$0.05$ & 4 & $\{2, 3, 4, 6, 8, 9\}$ \\
$0.054$ & 2 & $\{2, 3, 4, 5, 6, 7, 8, 9\}$ \\
$0.1$ & 1 & $\{1, \dots, 9\}$ \\
\bottomrule
\end{tabular}
\end{table}

\section{Conclusion}

In this work, we presented \textbf{\our}, a parameter-efficient continual learning method that dynamically adjusts model capacity to the complexity of the task stream. By revisiting the rigid ``one-adapter-per-task'' paradigm, we demonstrated that high performance does not require linear parameter growth. Instead, we introduced a differentiable gating mechanism that effectively acts as a structural switch, allowing the model to learn new parameters only when the semantic novelty of a task demands it.

Our empirical analysis on CIFAR-100, ImageNet-A, and ImageNet-R yields two key insights. First, the stability--plasticity dilemma can be mitigated by decoupling feature learning from structural allocation; our method preserves the plasticity of unconstrained LoRA while achieving stability through selective reuse. Second, standard CL benchmarks contain significant redundancy; we showed that up to 70\% of task-specific adapters can be pruned with no loss in accuracy and comparable forgetting.

We hope to motivate further theoretical study of when and why low-rank subspaces are shared across tasks.

\section*{Impact Statement}

Our work improves the efficiency of continual adaptation for foundation models by reducing unnecessary parameter growth. This can lower the computational and energy costs of maintaining large models over time, contributing to more sustainable machine learning practices.

As with other advances in model efficiency, reduced costs may enable wider deployment, which underscores the importance of responsible use and appropriate safeguards in downstream applications.

\bibliography{arxiv}

@inproceedings{DBLP:conf/iclr/DosovitskiyB0WZ21,
  author = {Dosovitskiy, Alexey and Beyer, Lucas and Kolesnikov, Alexander and Weissenborn, Dirk and Zhai, Xiaohua and Unterthiner, Thomas and Dehghani, Mostafa and Minderer, Matthias and Heigold, Georg and Gelly, Sylvain and Uszkoreit, Jakob and Houlsby, Neil},
  bibsource = {dblp computer science bibliography, https://dblp.org},
  booktitle = {9th International Conference on Learning Representations, {ICLR} 2021,
                  Virtual Event, Austria, May 3-7, 2021},
  publisher = {OpenReview.net},
  title = {An Image is Worth 16x16 Words: Transformers for Image Recognition
                  at Scale},
  url = {https://openreview.net/forum?id=YicbFdNTTy},
  year = 2021
}

@article{jang2016categorical,
  title={Categorical reparameterization with gumbel-softmax},
  author={Jang, Eric and Gu, Shixiang and Poole, Ben},
  journal={arXiv preprint arXiv:1611.01144},
  year={2016}
}

@misc{cifar,
    title = {Learning Multiple Layers of Features from Tiny Images},
    author = {Alex Krizhevsky},
    year={2009}
}

@misc{loshchilov2019decoupledweightdecayregularization,
      title={Decoupled Weight Decay Regularization}, 
      author={Ilya Loshchilov and Frank Hutter},
      year={2019},
      eprint={1711.05101},
      archivePrefix={arXiv},
      primaryClass={cs.LG},
      url={https://arxiv.org/abs/1711.05101}, 
}

@inproceedings{
wang2023orthogonal,
title={Orthogonal Subspace Learning for Language Model Continual Learning},
author={Xiao Wang and Tianze Chen and Qiming Ge and Han Xia and Rong Bao and Rui Zheng and Qi Zhang and Tao Gui and Xuanjing Huang},
booktitle={The 2023 Conference on Empirical Methods in Natural Language Processing},
year={2023},
url={https://openreview.net/forum?id=L7ZBpZZ8Va}
}

@inproceedings{
wu2025sdlora,
title={{SD}-Lo{RA}: Scalable Decoupled Low-Rank Adaptation for Class Incremental Learning},
author={Yichen Wu and Hongming Piao and Long-Kai Huang and Renzhen Wang and Wanhua Li and Hanspeter Pfister and Deyu Meng and Kede Ma and Ying Wei},
booktitle={The Thirteenth International Conference on Learning Representations},
year={2025},
url={https://openreview.net/forum?id=5U1rlpX68A}
}

@misc{hu2021loralowrankadaptationlarge,
      title={LoRA: Low-Rank Adaptation of Large Language Models}, 
      author={Edward J. Hu and Yelong Shen and Phillip Wallis and Zeyuan Allen-Zhu and Yuanzhi Li and Shean Wang and Lu Wang and Weizhu Chen},
      year={2021},
      eprint={2106.09685},
      archivePrefix={arXiv},
      primaryClass={cs.CL},
      url={https://arxiv.org/abs/2106.09685}, 
}

@article{hendrycks2021nae,
  title={Natural Adversarial Examples},
  author={Dan Hendrycks and Kevin Zhao and Steven Basart and Jacob Steinhardt and Dawn Song},
  journal={CVPR},
  year={2021}
}

@article{DBLP:journals/corr/BengioLC13,
  author       = {Yoshua Bengio and
                  Nicholas L{\'{e}}onard and
                  Aaron C. Courville},
  title        = {Estimating or Propagating Gradients Through Stochastic Neurons for
                  Conditional Computation},
  journal      = {CoRR},
  volume       = {abs/1308.3432},
  year         = {2013},
  url          = {http://arxiv.org/abs/1308.3432},
  eprinttype    = {arXiv},
  eprint       = {1308.3432},
  bibsource    = {dblp computer science bibliography, https://dblp.org}
}

@article{HADSELL20201028,
title = {Embracing Change: Continual Learning in Deep Neural Networks},
journal = {Trends in Cognitive Sciences},
volume = {24},
number = {12},
pages = {1028-1040},
year = {2020},
issn = {1364-6613},
doi = {https://doi.org/10.1016/j.tics.2020.09.004},
url = {https://www.sciencedirect.com/science/article/pii/S1364661320302199},
author = {Raia Hadsell and Dushyant Rao and Andrei A. Rusu and Razvan Pascanu}
}

@InProceedings{liang2024inflora,
  author = {Liang, Yan-Shuo and Li, Wu-Jun},
  title = {InfLoRA: Interference-Free Low-Rank Adaptation for Continual Learning},
  booktitle = {Proceedings of the IEEE/CVF Conference on Computer Vision and Pattern Recognition (CVPR)},
  month = {June},
  year = {2024},
  pages = {23638--23647}
}

@article{hendrycks2021many,
  title={The Many Faces of Robustness: A Critical Analysis of Out-of-Distribution Generalization},
  author={Dan Hendrycks and Steven Basart and Norman Mu and Saurav Kadavath and Frank Wang and Evan Dorundo and Rahul Desai and Tyler Zhu and Samyak Parajuli and Mike Guo and Dawn Song and Jacob Steinhardt and Justin Gilmer},
  journal={ICCV},
  year={2021}
}

@INPROCEEDINGS{imagenet,
  author={Deng, Jia and Dong, Wei and Socher, Richard and Li, Li-Jia and Kai Li and Li Fei-Fei},
  booktitle={2009 IEEE Conference on Computer Vision and Pattern Recognition}, 
  title={ImageNet: A large-scale hierarchical image database}, 
  year={2009},
  volume={},
  number={},
  pages={248-255},
  doi={10.1109/CVPR.2009.5206848}}
\bibliographystyle{icml2026}

\newpage
\appendix
\onecolumn
\section{Appendix}

\subsection{Limitations}

First, the final keep/discard decision depends on hyperparameters such as the sparsity weight, gating temperature, and the duration of Phase 2 optimization. These choices may require tuning across backbones or datasets.

Second, while our method enables task-level pruning, it implicitly assumes that redundant adapters are not uniquely required for future tasks. In other words, when a task-specific adapter is removed, we assume its contribution is already subsumed by earlier learned adapters. Although our empirical results suggest a strong redundancy structure and a stable pruning hierarchy (Appendix~\ref{sec:pruning_consistency}), this assumption may not hold in settings with highly compositional tasks, where later tasks could depend on features that are not fully captured by earlier adapters.

Finally, our selection phase introduces additional optimization steps per task (albeit lightweight), and we have not yet studied the strongest scaling regimes (e.g., longer streams or higher-resolution backbones).

\subsection{Pruning Consistency Analysis}
\label{sec:pruning_consistency}
We report the full pruning frequency analysis for ImageNet-A, where individual decisions are noisier due to the larger number of fine-grained tasks.

Nevertheless, a \emph{soft} importance hierarchy emerges.
Table~\ref{tab:pruning_freq_ina} reports the fraction of runs (across 43 $\lambda$ values in the active pruning range) in which each task is pruned. Tasks~5 and~8 are pruned in 88\% of configurations, and a clear gradient of importance exists in between. This confirms that the gating mechanism captures meaningful task-level redundancy even when individual pruning decisions are noisy.

\begin{table*}[ht]
\centering
\caption{\textbf{ImageNet-A Pruning Frequency.} The percentage of runs (out of 43 configurations with $\lambda \in [0.01, 0.04]$) in which specific tasks were pruned.}
\label{tab:pruning_freq_ina}

\setlength{\tabcolsep}{6pt}
\begin{tabular}{ll}
\toprule
Tasks & Prune Frequency \\
\midrule
0 & 0\% \\
2, 3 & 16--21\% \\
9, 11, 18 & 33--40\% \\
1, 4, 12, 13, 15 & 42--49\% \\
10, 14, 16, 17, 19 & 54--66\% \\
6, 7 & 65--72\% \\
5, 8 & 88\% \\
\bottomrule
\end{tabular}
\end{table*}

\begin{figure*}[h]
    \begin{center}
    \includegraphics[width=0.7\linewidth]{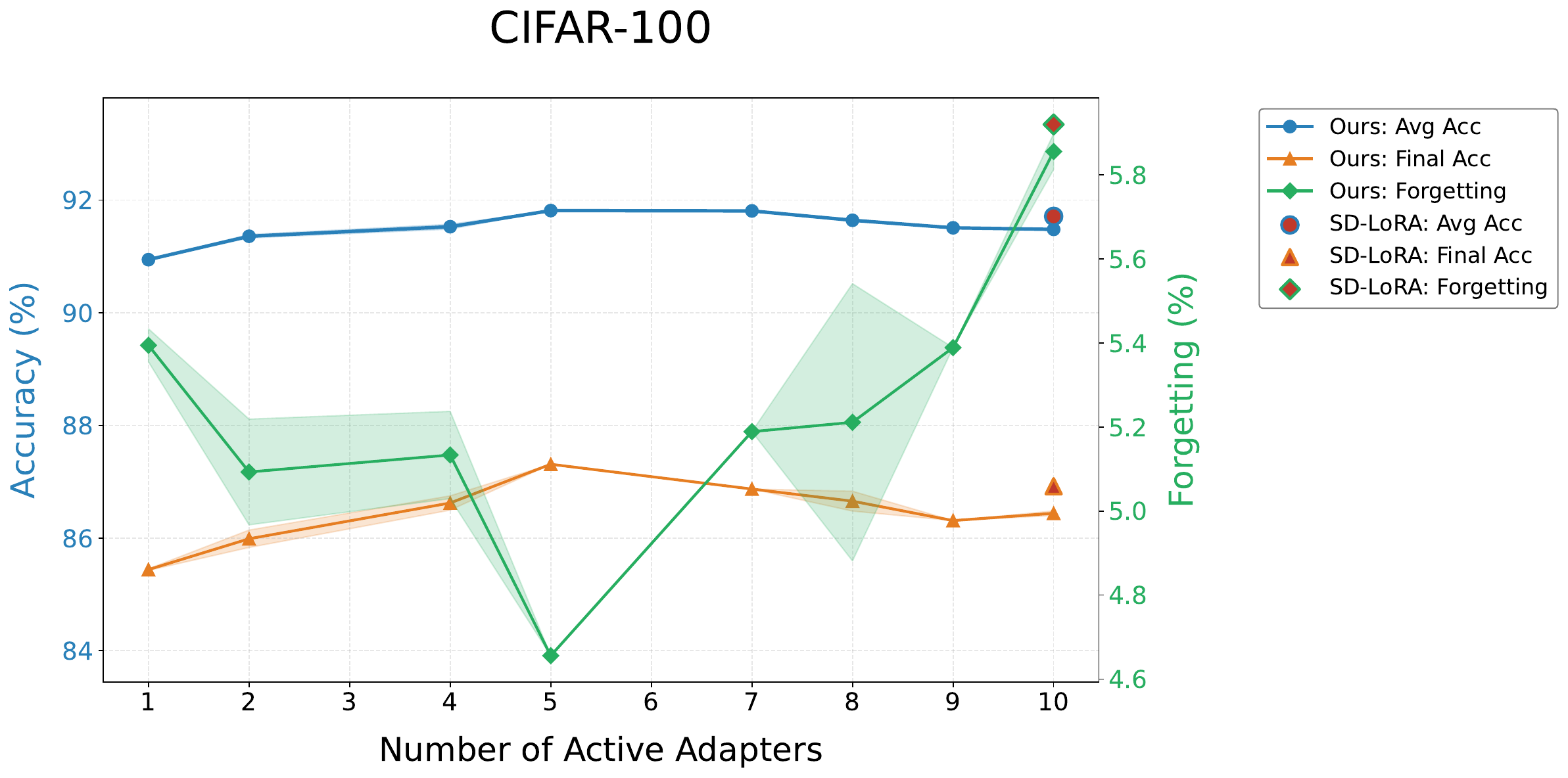}
    \caption{Sparsity--accuracy frontier on CIFAR-100 (order 1). We report average accuracy and forgetting as a function of the number of active adapters. Note that varying the regularization strength parameter $\lambda_\text{sparsity}$ implicitly changes the number of adapters, therefore it was not possible to test all points on the x axis.
    }
    \label{fig:sparsity_frontier}
    \end{center}
\end{figure*}

\subsection{Ablation Studies}
\label{sec:noise_ablation_appendix}

We study the impact of injecting stochastic Gumbel noise during Phase~1; by default we keep Phase~1 deterministic and add noise only in Phase~2. As shown in Table~\ref{tab:comparison_results}, adding noise in both phases does not improve and generally lowers average accuracy across all sparsity regimes, confirming that Phase~1 noise is unnecessary.

\begin{table}[H]
\centering
\small
\setlength{\tabcolsep}{8pt}
\renewcommand{\arraystretch}{1.2}
\caption{Gumbel-noise ablation on ImageNet-A (20 tasks) across different sparsity regimes. Results reported for Order 1.}
\label{tab:comparison_results}
\begin{tabular}{l l c c c c}
\toprule
Regime & Model variant & Avg.\ Acc.\ (\%)$\uparrow$ & Last Acc.\ (\%)$\uparrow$ & Forgetting (\%)$\downarrow$ & \#Ad$\downarrow$ \\
\midrule
\textbf{Dense} & Noise in both phases      & 64.93 & 53.46 & 17.56 & 20 \\
               & No phase 1 noise (Ours)   & \textbf{65.80} & 54.11 & 17.22 & 20 \\
               & No noise                  & 65.74 & \textbf{54.84} & \textbf{16.25} & 20 \\
\addlinespace[0.3em]
\textbf{Balanced} & Noise in both phases      & 62.01 & 50.89 & 18.37 & 11 \\
                  & No phase 1 noise (Ours)   & 62.19 & 51.22 & \textbf{15.89} & 10 \\
                  & No noise                  & \textbf{63.01} & \textbf{52.73} & 16.71 & \textbf{8} \\
\addlinespace[0.3em]
\textbf{Sparse} & Noise in both phases      & 61.69 & \textbf{54.64} & \textbf{13.54} & 3 \\
                & No phase 1 noise (Ours)   & \textbf{62.20} & 53.13 & 15.53 & 3 \\
                & No noise                  & 62.03 & 53.72 & 15.09 & 3 \\
\bottomrule
\end{tabular}
\end{table}

\subsection{Experiment Details}
\label{sec:hyperparams}

We provide the detailed hyperparameters used for our experiments on CIFAR-100, ImageNet-A,
and ImageNet-R in Table~\ref{tab:hyperparams}. The sparsity penalty
$\lambda_{\text{sparsity}}$ was selected via grid search on the first seed/ordering.

\begin{table}[H]
\centering
\caption{Hyperparameter settings for \our.}
\label{tab:hyperparams}
\small
\begin{tabular}{llll}
\toprule
\textbf{Parameter} & \textbf{CIFAR-100} & \textbf{ImageNet-A} & \textbf{ImageNet-R} \\
\midrule
Backbone        & ViT-B/16 (IN-21k) & ViT-B/16 (IN-21k) & ViT-B/16 (IN-21k) \\
LoRA Rank ($r$) & 10                & 10                & 10 \\
\# Tasks        & 10                & 20                & 20 \\
\midrule
\multicolumn{4}{c}{\textbf{Phase 1: Representation Learning}} \\
\midrule
Optimizer       & SGD               & Adam              & Adam \\
LR Scheduler    & Cosine            & Constant          & Constant \\
Learning Rate   & $8 \times 10^{-3}$ & $1 \times 10^{-2}$ & $1 \times 10^{-2}$ \\
Weight Decay    & $2 \times 10^{-4}$ & $2 \times 10^{-4}$ & $2 \times 10^{-4}$ \\
Epochs          & 20                & 20                & 20 \\
Batch Size      & 128               & 128               & 128 \\
Init $\alpha$   & 1.0               & 1.0               & 1.0 \\
Init Logit ($l_0$) & 0.5            & 0.5               & 0.5 \\
\midrule
\multicolumn{4}{c}{\textbf{Phase 2: Selection Learning}} \\
\midrule
Optimizer       & AdamW \citep{loshchilov2019decoupledweightdecayregularization}
                                    & AdamW             & AdamW \\
Learning Rate   & 0.05              & 0.05              & 0.05 \\
Epochs          & 1                 & 1                 & 1 \\
Batch Size      & 128               & 16                & 16 \\
$\lambda_{\text{sparsity}}$ (Range)
                & $[0.015,\,0.1]$   & $[0.01,\,0.04]$  & $[0.02,\,0.05]$ \\
Gumbel $\tau$   & 1.0               & 1.0               & 1.0 \\
\bottomrule
\end{tabular}
\end{table}

\paragraph{Combined Training Loss.}

For task $\mathcal{D}_t$, we optimize the network by combining cross-entropy loss with an $\ell_1$ sparsity regularization applied to the gates:

\begin{equation}
\label{eq:total_loss}
\mathcal{L} = \mathcal{L}_{\mathrm{CE}} + \lambda_{\text{sparsity}} \sum_{i=1}^{t} \beta_i ,
\end{equation}
where the sum runs over the per-task scalar gates and, since all previous gates are frozen, only $\beta_t$ receives a gradient.

\end{document}